\documentclass[11pt]{article}

\usepackage[final]{acl}

\usepackage{times}
\usepackage{latexsym}

\usepackage[T1]{fontenc}

\usepackage[utf8]{inputenc}

\usepackage{microtype}

\usepackage{inconsolata}

\usepackage{graphicx}
\usepackage{hyperref}
\usepackage{amsmath}
\usepackage{amsfonts} 
\title{mucAI at BAREC Shared Task 2025: Towards Uncertainty Aware Arabic Readability Assessment}

\author{Ahmed Abdou \\
  Independent Researcher. Munich, Germany \\
  \texttt{ahmedabdou1789@gmail.com}
  }

\begin{document}
\maketitle
\begin{abstract}
We present a simple, model-agnostic post-processing technique for fine-grained Arabic readability classification in the BAREC 2025 Shared Task (19 ordinal levels). Our method applies conformal prediction to generate prediction sets with coverage guarantees, then computes weighted averages using softmax-renormalized probabilities over the conformal sets.  This uncertainty-aware decoding improves Quadratic Weighted Kappa (QWK) by reducing high-penalty misclassifications to nearer levels. Our approach shows consistent QWK improvements of 1-3 points across different base models. In the strict track, our submission achieves QWK scores of 84.9\%(test) and 85.7\% (blind test) for sentence level, and 73.3\% for document level. For Arabic educational assessment, this enables human reviewers to focus on a handful of plausible levels, combining statistical guarantees with practical usability.
\end{abstract}


\section{Introduction}

Automatic readability assessment estimates how difficult a text will be for a target audience, a task essential for the design and advancement of pedagogically oriented NLP applications\citep{REAP, xia2019text}. In Arabic, this problem is particularly challenging due to morphological richness, and orthographic variation \cite{liberato2024strategies, benajiba2008arabic}. Recent work has advanced Arabic readability assessment through modeling and datasets \cite{saddiki2018feature, alhafni-etal-2024-samer, barec-corpus, habash-etal-2025-guidelines}. Most recently, the BAREC corpus \cite{barec-corpus} which offers 19 fine-grained levels. Nevertheless, even state-of-the-art models like AraBERT-v2 \cite{antoun2020arabert} remain prone to large-gap misclassifications and offer no principled means of quantifying prediction uncertainty. We address this by integrating conformal prediction \cite{vovk2005algorithmic} to produce statistically valid prediction sets and uncertainty-guided final predictions, reducing high-penalty errors and enabling compact, interpretable outputs for human-in-the-loop educational use. On the BAREC 2025 Shared Task, our method consistently improves QWK across base models, reaching $84.9\%$ on the test set and $85.7\%$ on the blind test at the sentence level, and $73.3\%$ on the blind test at the document level. Beyond leaderboard improvements, our method provides interpretable prediction sets and uncertainty estimates that enable more reliable readability assessment. Our implementation is open-sourced for reproducibility\footnote{\url{https://github.com/AhmedAbdel-Aal/mucAI-at-BAREC_2025}}.
\section{Background}

\subsection{Task and Data}
The BAREC Shared Task 2025 \cite{barec-task} targets fine-grained Arabic readability assessment across 19 ordered levels. The task builds on the BAREC corpus \cite{barec-corpus}, a manually annotated dataset containing over 69,000 sentences and more than one million words. The corpus provides mappings to multiple granularities (3, 5, and 7 readability levels); for detailed annotation guidelines, we refer readers to \cite{habash-etal-2025-guidelines}. We participated in both sentence-level and document-level variants of the strict track, where participants are restricted to using only the BAREC corpus for training. In the document-level task, a document's overall readability level is determined by its most difficult sentence. Given the ordinal nature of readability levels, the main evaluation metric is Quadratic Weighted Cohen's Kappa (QWK), which penalizes larger misclassifications more heavily \cite{cohen1968weighted}. This reflects the educational goal of avoiding assignments far from a student's level. We also report exact accuracy, adjacent accuracy (±1 of true label), Mean Absolute Error (MAE), and coarse-grained variants Acc7, Acc5, and Acc3, which collapse the 19 levels into 7, 5, and 3 bins. The shared task provides standard splits: training (54.8k), development (7.3k), test (7.3k), and blind test (3.4k), with the first three publicly available\footnote{\href{https://huggingface.co/collections/CAMeL-Lab/barec-shared-task-2025-6846ed8acb652c8d82aecd2a}{BAREC corpus shared-task-2025}}.

\subsection{Conformal Prediction}

Conformal Prediction (CP) \citep{vovk2005algorithmic, papadopoulos2002inductive} is a model-agnostic method that converts single predictions into prediction sets with statistical guarantees. Rather than predicting ``this text is Level 9'', CP produces ``this text is likely Level 7, 8, 9, 10, or 11''. Given a target miscoverage rate $\alpha$, CP guarantees that the true label appears in the prediction set with probability at least $1-\alpha$:
\begin{equation}
\mathbb{P}\big( Y \in C(X) \big) \geq 1 - \alpha
\end{equation}
where $C(X)$ is the predicted set for input $X$ and $Y$ is the true label. The method works by using a calibration set, data not seen during training, to learn how ``unusual'' different labels are for given inputs. This unusualness is captured by a nonconformity score $s(x,y)$: higher scores mean label $y$ is less plausible for input $x$ (more in appendix \ref{app:nonconf-scores}.). CP then sets a threshold $\hat{\tau}$ which is chosen as the $(1-\alpha)(n+1)$-quantile of these scores in the calibration set, ensuring the coverage guarantee. For any new input $x$, the prediction set includes all labels below this threshold:
\begin{equation}
C(x) = \{ y \in \mathcal{Y} : s(x,y) \leq \hat{\tau} \}
\end{equation}
\section{Method}

We use AraBERT-v2 \cite{antoun2020arabert} as the backbone, following the strongest BAREC baselines \cite{barec-corpus}. The original benchmark reports four preprocessing pipelines based on CAMeL tools \cite{obeid2020camel} (Word, Lex, D3Lex, D3Tok) but we could not run the CAMeL D3 analyzer in our environment. Because BAREC releases the dev/test sentences already preprocessed with these pipelines, we include them for comparison. For the blind split, however, only raw text is provided; we therefore adopt AraBERT’s recommended Farasa segmentation \cite{abdelali2016farasa}. For training objectives, we replicate the benchmark baselines: Cross-Entropy (CE) and Earth Mover’s Distance (EMD) \cite{EMD}, and an ordinal Regression variant. Our addition is a Focal-loss objective \cite{lin2017focal} tailored to the long-tailed 19-level label distribution; we report it alongside the baselines and simple ensembles: probability averaging, and majority voting.

Our post-processing approach combines conformal prediction with expected value decoding. We first generate prediction sets with coverage guarantees, then produce final predictions by averaging within these sets. We apply CP only to the probabilistic classifiers (CE/EMD/Focal); the Regression head is reported as point predictions only.

\paragraph{Prerequisites and Notation.}
Let $\mathcal{Y}=\{1 ,..., 19\}$ denote the ordered labels. A trained classifier produces posterior probabilities $p(y\mid x)$ for input $x$. For any $x$, we build form a conformal prediction set $C(x)\subseteq\mathcal{Y}$ and then decode to a single label.

\paragraph{Calibration and Tuning Protocol.}
We split the official development set into two stratified halves: a calibration split (\textit{dev-cal}) for learning conformal thresholds, and a tuning split (\textit{dev-tune}) for hyperparameter selection and evaluation. See the split details in Table \ref{dev-split} in Appendix \ref{app:dev-split}.

\paragraph{Set Construction.}
We evaluate three standard nonconformity score functions for multiclass conformal prediction: naïve (inverse-probability), APS (Adaptive Prediction Sets) \cite{APS}, and RAPS (Regularized APS) \cite{RAPS}.

\paragraph{Renormalization within the set.}
We first renormalize probabilities within the conformal set
\[
p_{C}(y\mid x)\;=\;\frac{p(y\mid x)}{\sum_{j\in C(x)} p(j\mid x)}\quad\text{for } y\in C(x).
\]
We then predict the rounded posterior mean
\[
\hat{y}(x)\;=\;\mathrm{round}\!\left(\sum_{y\in C(x)} y\, p_{C}(y\mid x)\right).
\]

The choice of weighted mean is motivated by its role as the Bayes-optimal point estimator under quadratic loss. While this is not strictly optimal for our discrete classification setting, we employ it as a computationally simple heuristic that aligns with the quadratic penalty structure of the primary evaluation metric (QWK). For the document-level track, we applied our best-performing sentence-level model to all sentences in a document and assigned the document’s readability as the maximum predicted level across its sentences, following the shared task definition. We report the full experimental setup in appendix \ref{app:experimental-setup}.

\section{Results}

Dev/test results demonstrate that clitic-aware preprocessing substantially improves performance: Farasa and D3Tok consistently outperform word-level and lexical baselines, with Farasa achieving the best QWK scores under CE, EMD, and regression losses, and on par under Focal loss. Given Farasa's consistent performance across dev/test splits and its availability as the only accessible preprocessor for blind evaluation, we standardize on Farasa preprocessing for all subsequent experiments (full results in Appendix ~\ref{app:full-tables}).

Table~\ref{test-results} reports sentence-level results on the BAREC 2025 test set. +CP improves QWK over each baseline while reducing exact Acc, and increases ±1Acc. The strongest single model is Focal+CP (QWK 84.4; +2.6 over Focal); CE+CP and EMD+CP gain +1.6 and +1.1 QWK, respectively. The Avg and Most Common ensembles also improve QWK (to 84.9 and 84.6) and reduce Dist (down to 1.01). To quantify headroom if a user could reliably choose from the CP set, we add a non-deployable Oracle: it selects the gold label whenever it lies in the CP set, otherwise falls back to Focal+CP. This upper bound reaches QWK 95.3 and Acc 94.8, closely tracking the target coverage ($\alpha$=0.10), and illustrates the potential of human-in-the-loop use of CP sets. Results on the blind test set (Table~\ref{blindtest-results}) validate the robustness of our approach. The ensemble averaging method achieves the highest performance at 85.7 QWK, while individual CP-enhanced models reach competitive scores of 84.3 (CE), 84.6 (EMD), and 85.3 (Focal). The regression baseline achieves 85.41 QWK, demonstrating strong performance of the regression formulation without post-processing. The consistent pattern of QWK improvements across different loss functions and evaluation sets demonstrates the generalizability of our conformal prediction approach.

\begin{table*}
  \centering
  \begin{tabular}{lccccccc}
    \hline
    \textbf{Model Variant} & \textbf{QWK} & \textbf{Acc$^{19}$} & \textbf{±1 Acc$^{19}$} & \textbf{Dist} & \textbf{Acc$^{3}$} & \textbf{Acc$^{5}$} & \textbf{Acc$^{7}$} \\
    \hline
    CE (Baseline) & 82.6 & \textbf{55.5} & 71.6 & 1.04 & 79.8 & 71.4 & \textbf{65.4} \\
    CE + CP & 84.3 & 50.3 & 72.9 & 1.03 & \textbf{80.1} & 70.1 & 63.8 \\
    \hline
    EMD (Baseline) & 82.8 & 54.4 & 71.4 & 1.04 & 79.7 & \textbf{71.5} & 64.6 \\
    EMD + CP & 83.9 & 49.4 & 73.4 & 1.04 & 79.7 & 70.4 & 63.3 \\
    \hline
    Focal (Baseline) & 81.8 & 55.4 & 71.7 & 1.07 & 79.7 & 71.4 & 65.3 \\

    Focal + CP & 84.4 & 42.7 & \textbf{74.5} & 1.08 & 78.0 & 67.9 & 61.0  \\
    \hline
    Regression (Baseline) & 83.8 & 42.0 & 73.2 & 1.12 & 78.0 & 67.3 & 59.8  \\
    \hline
     Average & \textbf{84.9} & 47.3 & 74.0 & 1.03 & 79.8 & 69.6 & 63.0 \\
    Most Common & 84.6 & 49.6 & 74.4 & \textbf{1.01} & \textbf{80.1} & 70.9 & 64.4\\
    \hline
     Oracle Decoder & 95.3 & 94.8 & 95.3 & 0.20 & 96.4 & 95.6 & 95.3 \\
    \hline

  \end{tabular}
  \caption{\label{test-results}
   BAREC test, sentence-level. ``Baseline'' = fine-tuned point decoder. ``+CP'' = conformal prediction ($\alpha$=0.10) with our QWK-aligned mean-in-set decoder; applied to CE/EMD/Focal only (Regression is point-only). ``Oracle'' = upper bound that selects the gold label if it lies in the CP set; otherwise falls back to Focal+CP. All results use Farasa preprocessing.
  }
\end{table*}

\begin{table}
  \centering
  \begin{tabular}{lc}
    \hline
    \textbf{Model Variant} & \textbf{QWK} \\
    \hline
    CE (Baseline) & 82.6 \\
    CE + CP & 84.3 \\
    EMD (Baseline) & - \\
    EMD + CP & 84.6 \\
    Focal (Baseline) & - \\
    Focal + CP & 85.3 \\
    Regression (Baseline) & 85.4 \\
    \hline
    Average & \textbf{85.7} \\
    Most Common & 84.8 \\
    \hline
    Document-level (Max over sentences) & 73.3 \\
    \hline
  \end{tabular}
  \caption{\label{blindtest-results}
    Blind test set QWK results. Missing baseline values (--) indicate models not submitted without CP enhancement. 
    Document-level results use the maximum predicted sentence-level difficulty per document.
  }
\end{table}
\section{Discussion}

We analyze our conformal prediction approach with the focal loss model and APS at $\alpha=0.1$, the best-performing setting on the dev-tune split. The analysis highlights two aspects: (1) coverage reliability and failure patterns, (2) error redistribution underlying improvements in ordinal metrics.

\subsection{CP Coverage Analysis}
\begin{figure}[t]
  \includegraphics[width=\columnwidth]{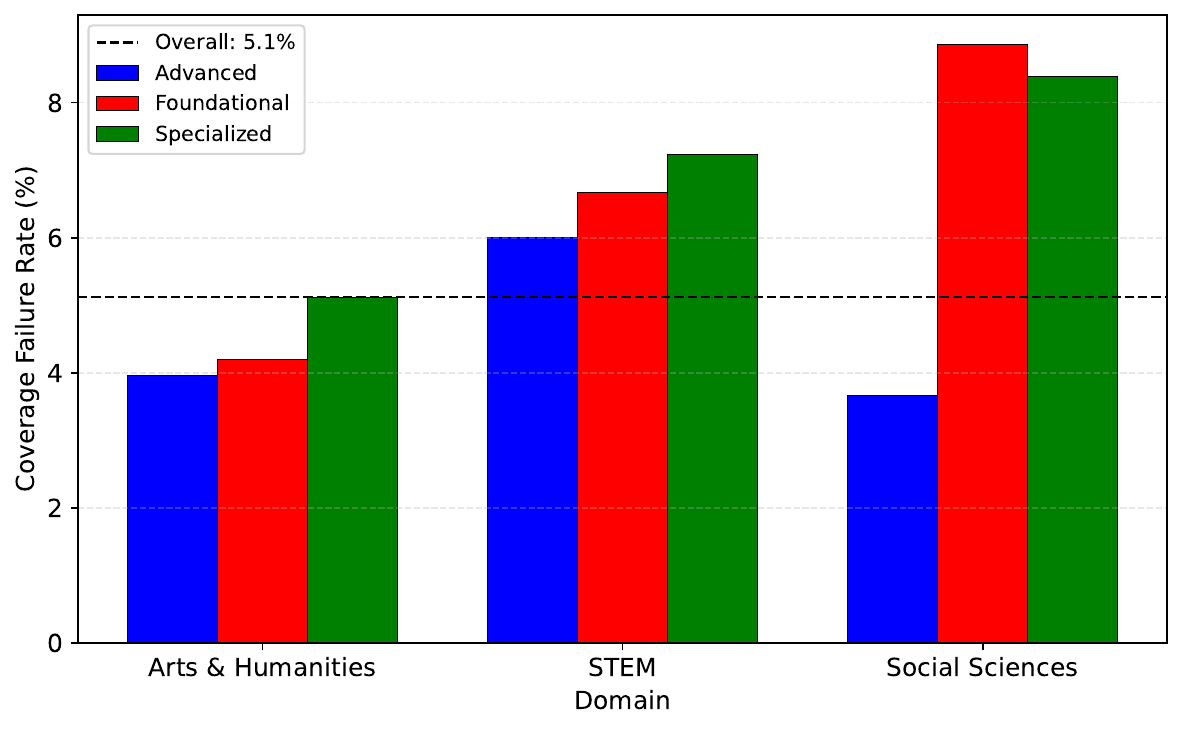}
\caption{Coverage failure rates by domain and text class. Each domain shows three grouped bars representing Advanced, Foundational, and Specialized text classes. The dashed line shows the overall failure rate (5.12\%).}
\label{fig:domain_plot_failure}
\end{figure}

Using $\alpha = 0.1$ targeting 90\% coverage, we report 94.88\% empirical coverage with an average set size of 5 levels, a substantial reduction from the full 19-class space. This means that in nearly 95\% of Arabic texts, the correct readability level appears in a compact, interpretable set. The remaining 5.12\% coverage failures show systematic domain variation: 4.3\% for Arts \& Humanities (70/1,625), 6.1\% for STEM (10/163), and 7.1\% for Social Sciences (38/535). We define failure rate as the proportion of cases where the true label falls outside the conformal prediction set. Figure~\ref{fig:domain_plot_failure} reveals that failures are not uniformly distributed across text types. Social Sciences exhibits the highest rates, particularly for Foundational and Specialized texts (8-9\% failure rates), while Arts \& Humanities remains close to the overall rate. STEM shows elevated failure rates (6-7\%) across all text classes. This variation suggests that domain-adaptive calibration strategies could improve coverage reliability for challenging text types. Additional coverage diagnostics are provided in Appendix \ref{app:cp-cov-plots}.

\subsection{Why QWK improves despite lower exact accuracy}
QWK increases because many large errors shrink while only a smaller set of perfect predictions become near misses.
On the dev–tune split, CP turned $362$ perfect predictions into errors ($15.6\%$), and $86.7\%$ of these new errors were only $\pm 1$ level.
At the same time, $397$ originally incorrect predictions improved ($17.1\%$): $80.6\%$ shrank by $1$ level, $14.7\%$ by $2$, $3.1\%$ by $3$, and $1.6\%$ by $4$. Since QWK penalizes errors by the \emph{squared} distance, shrinking many large mistakes yields big gains
(e.g., reducing a $4$-level error to $1$ cuts the penalty from $16$ to $1$).

\section{Conclusions and Future Work}
We presented a simple, model-agnostic post-processing method for Arabic readability assessment that combines conformal prediction with expected value decoding. Applied to the BAREC Shared Task 2025, our approach achieved consistent QWK gains of 1-3 points across multiple base models. In the strict track, our submission achieves QWK scores of 84.9\% (test) and 85.7\% (blind test) for sentence level, and 73.3\% for document level. 
Beyond leaderboard gains, the method produces compact prediction sets with statistical coverage guarantees, offering both improved accuracy and interpretable outputs for human-in-the-loop use.

Future work could extend this approach in several ways. 
Mondrian conformal prediction could calibrate separately for different text types or complexity ranges, potentially reducing coverage failures in difficult cases. Multi-granularity training using the BAREC mappings (3-, 5-, and 7-level schemes) may improve generalization across difficulty levels. Finally, rule-based or heuristic decoding strategies informed by the official annotation guidelines \cite{habash-etal-2025-guidelines} could refine label selection from CP sets by leveraging linguistic cues and common annotation patterns.

\section{Limitations}
While our approach improves QWK and reduces high-penalty errors, several limitations remain. Most error reductions occur within medium difficulty ranges, leaving large-gap errors at higher levels (e.g., 15--19) largely unresolved. The effectiveness of our approach depends on the base model’s calibration: overconfident but incorrect probability estimates can lead to suboptimal conformal sets, and renormalization may not fully correct such biases. Finally, our CP implementation yields slightly conservative coverage (94\% vs.\ 90\% target), suggesting room for tighter calibration or adaptive thresholding.

\bibliography{custom}

\appendix

\section{Appendix A}
\label{sec:appendix}

\subsection{Nonconformity Scores}
\label{app:nonconf-scores}
In conformal prediction, a \emph{nonconformity score} $s(x,y)$ quantifies how atypical a candidate label $y$ is for an instance $x$ given the model's output distribution $p(y\mid x)$. We evaluate three standard multiclass scoring functions:

\paragraph{Na\"ive (Inverse Probability).}
The simplest approach uses the complement of the predicted probability:
\begin{equation}
s_{\text{naive}}(x,y) = 1 - p(y \mid x)
\end{equation}
This yields smaller scores for high-probability labels, producing larger prediction sets for low-confidence predictions.

\paragraph{Adaptive Prediction Sets (APS)} \cite{APS}.
Let $\pi_1, \pi_2, \ldots, \pi_K$ denote the classes sorted in descending order of their probabilities $p(\pi_1 \mid x) \geq p(\pi_2 \mid x) \geq \cdots \geq p(\pi_K \mid x)$. For a given label $y$, let $r(y)$ be its rank in this sorted order. The APS score is the cumulative probability mass up to and including label $y$:
\begin{equation}
s_{\text{aps}}(x,y) = \sum_{j=1}^{r(y)} p(\pi_j \mid x)
\end{equation}

\paragraph{Regularized Adaptive Prediction Sets (RAPS)} \cite{RAPS}.
RAPS extends APS by adding a linear rank-based penalty:
\begin{equation}
s_{\text{raps}}(x,y) = \sum_{j=1}^{r(y)} p(\pi_j \mid x) + \lambda \cdot r(y)
\end{equation}
where $\lambda \geq 0$ is the regularization parameter controlling the size-coverage trade-off. In this work, we set $\lambda = 0.01$.

\subsection{Experimental Setup}
\label{app:experimental-setup}
All experiments were conducted on a single NVIDIA A100 GPU using Google Colab Pro. Training was performed for 6 epochs with a batch size of 64, a learning rate of $5\times 10^{-5}$, and the Adam optimizer. The best checkpoint was selected based on development set performance measured by Quadratic Weighted Kappa (QWK).

\subsection{CP Coverage Plots}
\label{app:cp-cov-plots}
\begin{figure}[t]
  \includegraphics[width=\columnwidth]{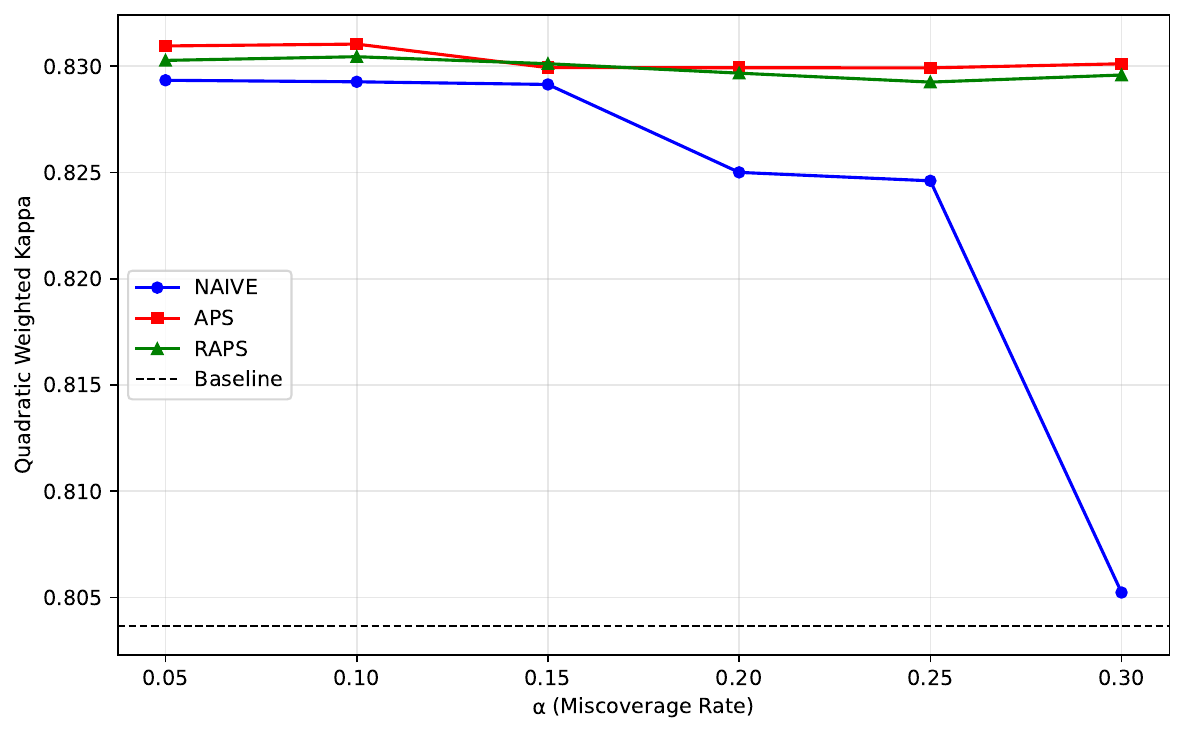}
\caption{Quadratic Weighted Kappa performance vs. miscoverage rate ($\alpha$) for three conformal prediction scoring methods on the dev-tune split. The dashed line represents baseline performance without conformal prediction.}  \label{fig:plot_1}
\end{figure}

\begin{figure}[t]
  \includegraphics[width=\columnwidth]{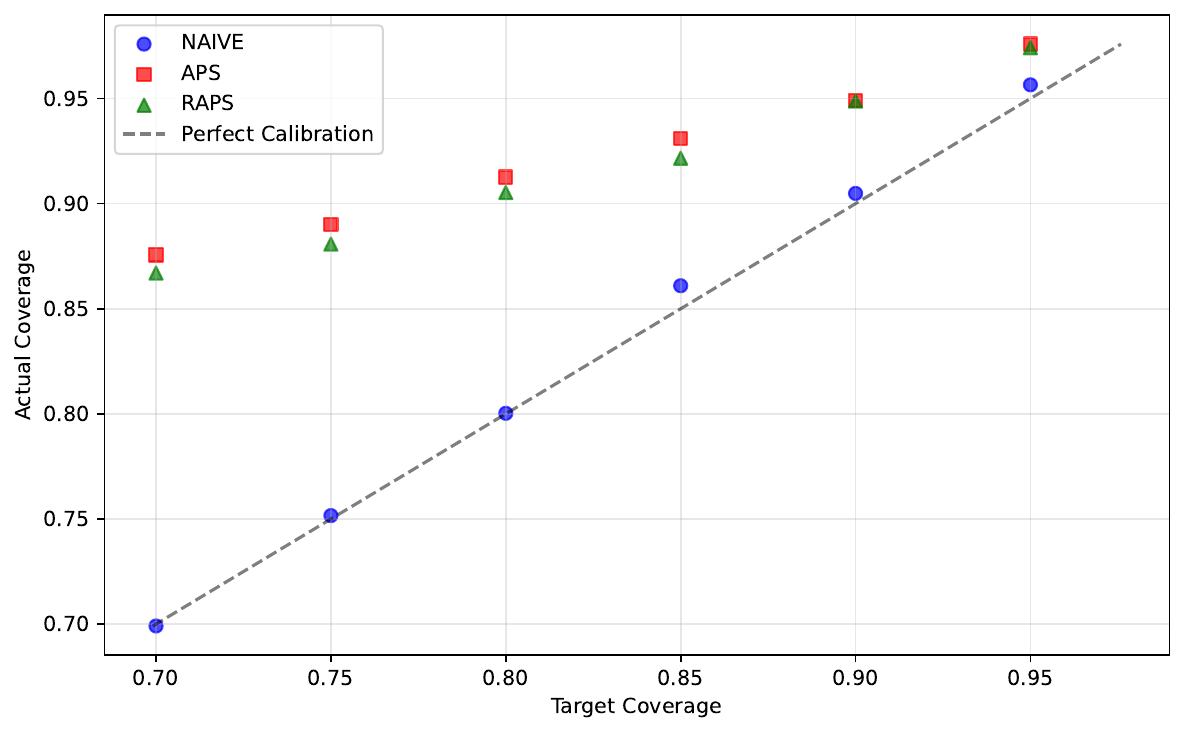}
\caption{Coverage calibration quality showing actual vs. target coverage rates. The dashed line represents perfect calibration where actual coverage equals target coverage.}  \label{fig:plot_2}
\end{figure}

\begin{figure}[t]
  \includegraphics[width=\columnwidth]{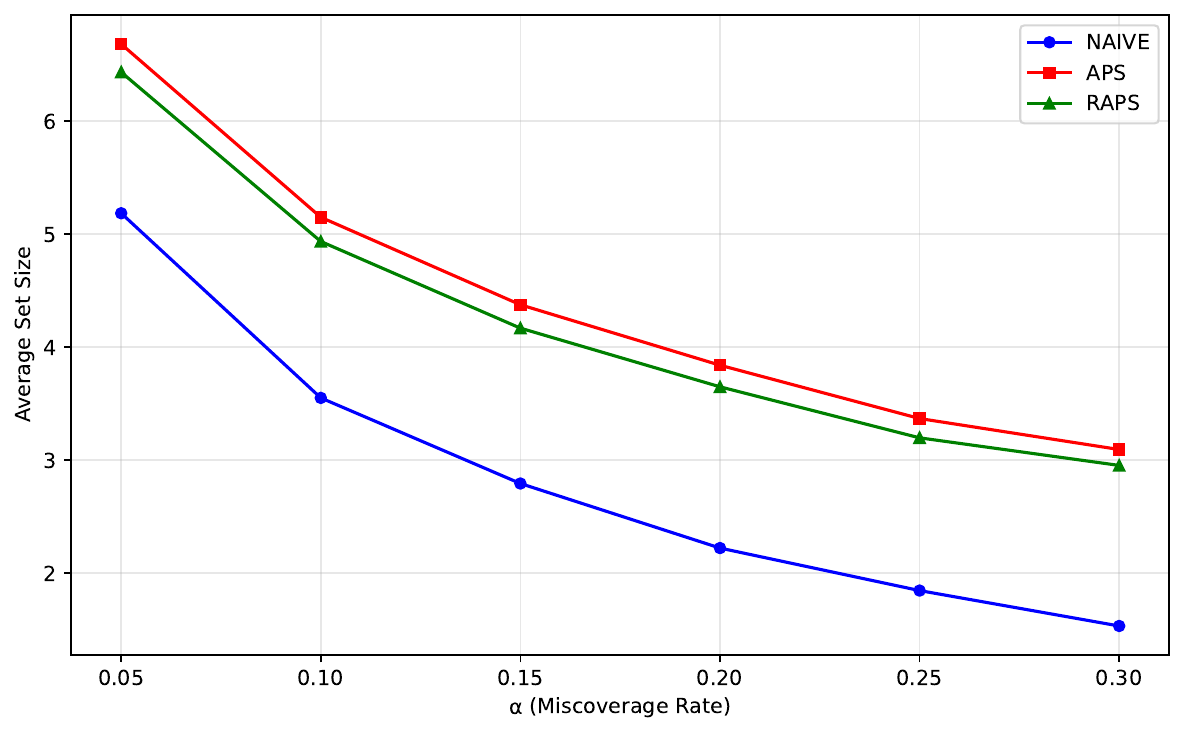}
\caption{Average prediction set sizes across miscoverage rate ($\alpha$) for the three conformal prediction scoring methods.}  \label{fig:plot_3}
\end{figure}

To better understand the behavior of our conformal prediction variants, we provide supplementary plots analyzing performance, coverage calibration, and set size trends across different miscoverage rates $\alpha$. In Figure \ref{fig:plot_1}, we show the relationship between miscoverage rate $\alpha$ and Quadratic Weighted Kappa (QWK) for three conformal prediction methods on the dev-tune set. APS and RAPS maintain stable QWK across all $\alpha$ values, consistently outperforming the baseline. The naïve method degrades sharply beyond $\alpha$ > 0.2, indicating poor robustness when allowing larger miscoverage.
In Figure \ref{fig:plot_2}, we plot the actual coverage against the target coverage for Naïve, APS, and RAPS methods. All methods achieve coverage above the target across the range, indicating slight conservativeness. This effect is most pronounced for APS, which consistently overshoots the target coverage. Such conservative calibration ensures statistical validity but may produce larger prediction sets than necessary, potentially impacting their interpretability. Finally, figure \ref{fig:plot_3} shows the relationship between the miscoverage rate $\alpha$ and the average prediction set size for the three nonconformity scoring methods. For $\alpha$, APS and RAPS yield larger sets than the naïve method, with APS producing the widest sets.

\subsection{Preprocessing \& Loss Ablations}
\label{app:full-tables}
\begin{table*}
  \centering
  \begin{tabular}{lccccc}
    \hline
    \textbf{Loss} & \textbf{Input} & \textbf{QWK} & \textbf{Acc$^{19}$} & \textbf{±1 Acc$^{19}$} & \textbf{Dist} \\
    \hline
    CE & Word & 77.6 & 53.4 & 68.2 & 1.24 \\
    CE & Lex & 76.4 & 49.0 & 66.1 & 1.32 \\
    CE & D3Lex & 79.8 & 53.0 & 68.3 & 1.19 \\
    CE & D3Tok & 81.4 & 53.3 & \textbf{70.9} & 1.14 \\
    CE & Farasa & \textbf{80.2} & \textbf{55.5} & 70.6 & \textbf{1.13} \\
    \hline
    EMD & Word & 78.2 & 52.0 & 67.3 & 1.24 \\
    EMD & Lex & 79.5 & 48.8 & 66.8 & 1.24 \\
    EMD & D3Lex & 80.4 & 52.2 & 68.3 & 1.18 \\
    EMD & D3Tok & 81.2 & 53.1 & 69.8 & 1.13 \\
    EMD & Farasa & \textbf{81.4} & \textbf{54.8} & \textbf{71.0} & \textbf{1.10} \\
    \hline
    Regression & Word & 79.3 & 38.5 & 69.4 & 1.30 \\
    Regression & Lex & 80.9 & 35.8 & 69.2 & 1.31 \\
    Regression & D3Lex & 82.3 & 38.7 & 70.7 & 1.26 \\
    Regression & D3Tok & 82.4 & 40.7 & 71.5 & 1.20 \\
    Regression & Farasa & \textbf{82.9} & \textbf{43.3} & \textbf{72.5} & \textbf{1.15} \\
    \hline
    Focal & Word & 77.6 & 52.6 & 67.6 & 1.25 \\
    Focal & Lex & 77.9 & 49.4 & 67.0 & 1.27 \\
    Focal & D3Lex & 80.0 & 53.4 & 69.1 & 1.18 \\
    Focal & D3Tok & \textbf{80.5} & 56.0 & \textbf{71.1} & \textbf{1.12} \\
    Focal & Farasa & 80.4 & \textbf{56.1} & 71.0 & \textbf{1.12} \\
    \hline
  \end{tabular}
  \caption{\label{abaltion-dev-results}
   AraBERTv2 results on the BAREC Development set across different loss functions and input representations.
  }
\end{table*}

\begin{table*}
  \centering
  \begin{tabular}{lccccc}
    \hline
    \textbf{Loss} & \textbf{Input} & \textbf{QWK} & \textbf{Acc$^{19}$} & \textbf{±1 Acc$^{19}$} & \textbf{Dist} \\
    \hline
    CE & Word & 79.2 & 54.0 & 68.6 & 1.17 \\
    CE & Lex & 78.4 & 49.7 & 66.9 & 1.23 \\
    CE & D3Lex & 80.6 & 53.2 & 68.1 & 1.14 \\
    CE & D3Tok & 81.9 & 52.8 & 70.9 & 1.10 \\
    CE & Farasa & \textbf{82.6} & \textbf{55.5} & \textbf{71.6} & \textbf{1.04} \\
    \hline
    EMD & Word & 80.7 & 53.3 & 68.9 & 1.13 \\
    EMD & Lex & 80.6 & 49.6 & 67.0 & 1.18 \\
    EMD & D3Lex & 81.3 & 53.3 & 69.6 & 1.11 \\
    EMD & D3Tok & 81.7 & 52.7 & 69.3 & 1.10 \\
    EMD & Farasa & \textbf{82.8} & \textbf{54.5} & \textbf{71.4} & \textbf{1.04} \\
    \hline
    Regression & Word & 81.4 & 38.8 & 70.4 & 1.23 \\
    Regression & Lex & 81.4 & 35.5 & 70.1 & 1.26 \\
    Regression & D3Lex & 82.8 & 39.2 & 70.9 & 1.18 \\
    Regression & D3Tok & 83.1 & 40.7 & 72.2 & 1.15 \\
    Regression & Farasa & \textbf{83.8} & \textbf{42.0} & \textbf{73.2} & \textbf{1.11} \\
    \hline
    Focal & Word & 79.9 & 53.9 & 69.4 & 1.14 \\
    Focal & Lex & 79.5 & 50.6 & 67.7 & 1.19 \\
    Focal & D3Lex & 80.9 & 53.1 & 69.6 & 1.13 \\
    Focal & D3Tok & \textbf{82.2} & 55.2 & 71.2 & \textbf{1.06} \\
    Focal & Farasa & 81.8 & \textbf{55.4} & \textbf{71.7} & 1.07 \\
    \hline
  \end{tabular}
  \caption{\label{abaltion-test-results-appendix}
   AraBERTv2 results on the BAREC Test set across different loss functions and input representations.
  }
\end{table*}

\subsection{Dev Data Split}
\label{app:dev-split}

\begin{table*}
  \centering
  \begin{tabular}{cccccc}
    \hline
    \textbf{Class} & \textbf{Original} & \textbf{\%} & \textbf{Dev-Cal} & \textbf{Dev-Tune} & \textbf{Split Ratio} \\
    \hline
    1 & 44 & 0.6 & 32 & 12 & 73:27 \\
    2 & 68 & 0.9 & 49 & 19 & 72:28 \\
    3 & 182 & 2.5 & 126 & 56 & 69:31 \\
    4 & 78 & 1.1 & 55 & 23 & 71:29 \\
    5 & 417 & 5.7 & 284 & 133 & 68:32 \\
    6 & 189 & 2.6 & 130 & 59 & 69:31 \\
    7 & 701 & 9.6 & 476 & 225 & 68:32 \\
    8 & 613 & 8.4 & 417 & 196 & 68:32 \\
    9 & 236 & 3.2 & 162 & 74 & 69:31 \\
    10 & 1012 & 13.8 & 686 & 326 & 68:32 \\
    11 & 409 & 5.6 & 279 & 130 & 68:32 \\
    12 & 1491 & 20.4 & 1010 & 481 & 68:32 \\
    13 & 349 & 4.8 & 239 & 110 & 68:32 \\
    14 & 1072 & 14.7 & 727 & 345 & 68:32 \\
    15 & 258 & 3.5 & 177 & 81 & 69:31 \\
    16 & 114 & 1.6 & 80 & 34 & 70:30 \\
    17 & 49 & 0.7 & 36 & 13 & 73:27 \\
    18 & 13 & 0.2 & 10 & 3 & 77:23 \\
    19 & 15 & 0.2 & 12 & 3 & 80:20 \\
    \hline
    \textbf{Total} & \textbf{7310} & \textbf{100.0} & \textbf{4981} & \textbf{2329} & \textbf{68:32} \\
  \end{tabular}
  \caption{\label{dev-split}
   Development set stratified split into calibration (Dev-Cal) and tuning (Dev-Tune) subsets.
  }
\end{table*}

\end{document}